\documentclass[12pt]{article}
\usepackage{amsthm,amsmath,amssymb,bbm,bm}
\usepackage{multirow}
\usepackage{xr-hyper}
\usepackage[pdftex]{graphicx}
\usepackage{subfigure}
\usepackage{wrapfig}
\usepackage{CJK}
\usepackage{array}
\usepackage{url}
\usepackage{booktabs}
\usepackage{algorithm}
\usepackage[dvipsnames]{xcolor}
\usepackage{algorithmic}
\usepackage{placeins}
\usepackage{mathtools}
\usepackage{romannum}
\usepackage{mathrsfs}
\usepackage{dsfont}
\usepackage{natbib}
\usepackage{relsize}
\usepackage{rotating}
\usepackage{enumitem}
\usepackage{setspace}
\usepackage{tikz}
\usepackage{multirow}
\usepackage{tikz}
\usepackage{caption}
\usepackage{subcaption}
\usetikzlibrary{shapes,shapes.multipart}
\usetikzlibrary{arrows}
\usepackage[page,header]{appendix}
\usepackage{titletoc}
\usepackage{titling}

\usepackage{multibib}

\newcites{SM}{References}

\usepackage[colorlinks]{hyperref}
\hypersetup{
    colorlinks,
    linkcolor={red},
    filecolor={red}, 
    urlcolor={blue},
    citecolor={blue}
}

\makeatletter
\newcommand*{\addFileDependency}[1]{
  \typeout{(#1)}
  \@addtofilelist{#1}
  \IfFileExists{#1}{}{\typeout{No file #1.}}
}
\makeatother
\newcommand{\noindep}{\not\!\perp\!\!\!\perp}

\newcommand{\indep}{\perp \!\!\! \perp}

\usepackage{etoolbox}
\newcommand{\zerodisplayskips}{%
  \setlength{\abovedisplayskip}{6.5pt}%
  \setlength{\belowdisplayskip}{6.5pt}%
  \setlength{\abovedisplayshortskip}{6.5pt}%
  \setlength{\belowdisplayshortskip}{6.5pt}}
\appto{\normalsize}{\zerodisplayskips}
\appto{\small}{\zerodisplayskips}
\appto{\footnotesize}{\zerodisplayskips}

\newcommand{\yl}[1]{\textcolor{blue}{#1}}

\makeatletter
\newcommand{\vast}{\bBigg@{3}}
\makeatother
\newtheorem{asu}{Assumption}
\usepackage{chngcntr}
\counterwithin{equation}{section}
\definecolor{ao(english)}{rgb}{0.0, 0.5, 0.0}

\theoremstyle{plain}

\newtheorem{thm}{Theorem}

\counterwithin{thm}{section}
\newtheorem{proposition}{Proposition}[thm]
\counterwithin{proposition}{section}

\counterwithin{corollary}{section}

\counterwithin{theorem}{section}

\def\##1\#{\begin{align}#1\end{align}}
\def\$#1\${\begin{align*}#1\end{align*}}

\def\beq#1\eeq{\begin{equation}#1\end{equation}}
\def\baa#1\eaa{\begin{eqnarray}#1\end{eqnarray}}
\def\bal#1\eal{\begin{align}#1\end{align}}

\DeclareMathOperator*{\argmin}{arg\,min}
\DeclareMathOperator*{\argmax}{arg\,max}

\newcommand{\blind}{1}

\usepackage[margin=0.9in]{geometry}
\begin{document}

\setcounter{page}{1}
\pagenumbering{arabic}


\if1\blind
{ \title{\bf Reinforcement Learning with Continuous Actions Under Unmeasured Confounding}

\author{
Yuhan Li\thanks{Department of Statistics, University of Illinois at Urbana-Champaign, Champaign, IL.}, \quad
Eugene Han\footnotemark[1], \quad
Yifan Hu\thanks{Department of Human Development and Family Studies, University of Illinois at Urbana-Champaign, Champaign, IL.}, \quad
Wenzhuo Zhou\thanks{Department of Statistics, University of California Irvine, Irvine, CA.},\\
Zhengling Qi\thanks{Department of Decision Sciences, The George Washington University, Washington, DC.}, \quad
Yifan Cui\thanks{School of Management \& Center for Data Science, Zhejiang University, Hangzhou, China.}, \quad
Ruoqing Zhu\footnotemark[1]
\thanks{Correspondence to Ruoqing Zhu \href{mailto:rqzhu@illinois.edu}{rqzhu@illinois.edu}}
}\fi

\if0\blind
{
    \title{\bf Reinforcement Learning with Continuous Actions Under Unmeasured Confounding}
} \fi

\newpage

\date{}

\maketitle
\vspace{-10mm}
\begin{abstract}
This paper addresses the challenge of offline policy learning in reinforcement learning with continuous action spaces when unmeasured confounders are present. While most existing research focuses on policy evaluation within partially observable Markov decision processes (POMDPs) and assumes discrete action spaces, we advance this field by establishing a novel identification result to enable the nonparametric estimation of policy value for a given target policy under an infinite-horizon framework. Leveraging this identification, we develop a minimax estimator and introduce a policy-gradient-based algorithm to identify the in-class optimal policy that maximizes the estimated policy value. Furthermore, we provide theoretical results regarding the consistency, finite-sample error bound, and regret bound of the resulting optimal policy. Extensive simulations and a real-world application using the German Family Panel data demonstrate the effectiveness of our proposed methodology.
\end{abstract}
\vfill

\newpage

\section{Introduction}

In practical applications of reinforcement learning (RL) \citep{sutton2018reinforcement}, the evaluation and optimization of policies using only pre-collected datasets has become essential. This need arises from the potential costs and safety risks associated with frequent interactions with the environment, such as real-world testing in autonomous driving \citep{zhu2020safe} and treatment selection in precision medicine \citep{luckett2019estimating,zhou2024policy}. As a result, there has been a growing interest in offline RL \citep{precup2000eligibility,levine2020offline}, which focuses on policy evaluation and optimization without requiring additional environment interactions.

Recent years have seen significant advancements in both off-policy evaluation (OPE) and off-policy learning (OPL). In OPE, popular methodologies include value-based approaches \citep{le2019batch,liao2021off}, importance sampling techniques \citep{liu2018breaking,xie2019towards}, and doubly robust methods that integrate value-based and importance sampling estimators \citep{uehara2020minimax,kallus2022doubly}. Additionally, confidence intervals for a target policy have been developed using bootstrapping \citep{hao2021bootstrapping}, asymptotic properties \citep{shi2022statistical}, and finite-sample error bounds \citep{zhou2023distributional}. In OPL, various algorithms have been proposed, demonstrating notable success across both discrete \citep{mnih2015human,luckett2019estimating,zhou2024estimating} and continuous action spaces \citep{kumar2020conservative,fujimoto2021minimalist,li2023quasi,zhou2024bi}.

A common assumption in the works mentioned above is the absence of unmeasured confounders. Specifically, these studies assume that all state variables are fully observed, with no unmeasured variables that might confound the observed actions. However, this assumption is unverifiable with offline data and is often violated in practical settings. Real-world instances include the influence of genetic factors in personalized medicine \citep{frohlich2018hype} and the complexities of path planning in robotics \citep{zhang2020causal}. To address these challenges, a recent line of research has focused on OPE within the framework of a confounded partially observable Markov Decision Process (POMDP), where the behavior policy generating the batch data may depend on unobserved state variables \citep{zhang2016markov}.

Various methods have been developed to identify the true policy value with the presence of unmeasured confounders. Within the contextual bandit setting (i.e., one decision point), causal inference frameworks have been established to identify the  treatment effect by sensitivity analysis \citep{bonvini2022sensitivity}, and through the use of instrumental variables \citep{cui2021semiparametric} and negative controls \citep{miao2018identifying,cui2023semiparametric}.  Existing methods for extensions to multiple decision points or the infinite horizon setting fall into two main categories. The first type considers i.i.d. confounders in the dynamic system and thus preserves the Markovian property \citep{zhang2016markov}. Under this framework, OPE methods have been formulated under various identification conditions, such as partial identification via sensitivity analysis \citep{kallus2020confounding, bruns2021model} and approaches that utilize instrumental variables or mediators \citep{li2021causal, shi2022off}. The second category explores the estimation of policy values in more general confounded POMDP models, where the Markovian assumption does not hold. These methods span a wide array of strategies, including the use of proxy variables \citep{ shi2022minimax, miao2022off,uehara2024future} or instrumental variables \citep{fu2022offline,xu2023instrumental}, spectral methods for undercomplete POMDPs \citep{ jin2020sample}, and techniques focusing on predictive state representation \citep{ cai2022reinforcement,guo2022provably}.

However, there are two less investigated issues in confounded POMDP settings. Firstly, existing methods mainly focus on discrete action spaces \citep{miao2022off,shi2022minimax,bennett2023proximal} despite many real-world scenarios requiring decision making on a continuous action space \citep{lillicrap2015continuous}.  A straightforward workaround in adapting existing methods to continuous domains is to discretize the continuous action space. However, this approach either introduces significant bias when using coarse discretization \citep{lee2018deep,cai2021jump} or encounters the the curse of dimensionality when applied to fine grids \citep{chou2017improving}. Secondly, the challenge of learning an optimal policy from a batch dataset, particularly in the presence of unmeasured confounders, is paramount in various fields, including personalized medicine \citep{lu2022pessimism} and robotics \citep{brunke2022safe}. A majority of existing methods, however, focus on policy evaluation rather than optimization. There are some recent efforts to address this gap. For instance, \cite{qi2023proximal} explore policy learning under the contextual bandit setting, and \cite{hong2023policy} investigate policy gradient methods under finite-horizon confounded POMDPs. In terms of the infinite-horizon, \cite{kallus2021minimax,fu2022offline} consider a restrictive memoryless confounders setting where the Markovian property is preserved, while \cite{guo2022provably,lu2022pessimism} focus on theoretical properties of the induced estimator under more general confounded POMDP settings, yet practical computational algorithms remain elusive. Thus, the development of computationally viable policy learning algorithms under infinite-horizon confounded MDPs, where the Markovian property is violated, remains a significant challenge.

Motivated by these, in this paper, we study the policy learning with continuous actions for confounded POMDPs over an infinite horizon. Our main contribution to the literature is threefold. First, relying on some time-dependent proxy variables, we extend the proximal causal inference framework \citep{miao2018identifying,tchetgen2020introduction} to infinite horizon setting, and establish a nonparametric identification result for OPE using $V$-bridge functions with continuous actions for confounded POMDPs. Leveraging the identification result, we propose an unbiased minimax estimator for the $V$-bridge function and introduce a computationally efficient Fitted-Q Evaluation (FQE)-type algorithm for estimating the policy value. Second, we develop a novel policy gradient algorithm that searches the optimal policy within a specified policy class by maximizing the estimated policy value. The proposed algorithm is tailored for continuous action spaces and provides enhanced interpretability of the optimal policy. Third, we thoroughly investigate the theoretical properties of the proposed methods on both policy evaluation and policy learning, including the estimator consistency, finite-sample bound for performance error, and sub-optimality of the induced optimal policy.  We validate our proposed method through extensive numerical experiments and apply it to the German Family Panel (Pairfam) dataset \citep{bruderl2023german}, where we aim at identifying  optimal strategies to enhance long-term relationship satisfaction.







\section{Preliminaries}
We consider an infinite-horizon confounded Partially Observable Markov Decision Process (POMDP) defined as $\mathcal{M}=\{\mathcal{S},\mathcal{O},\mathcal{A}, \mathbf{P},R,\gamma\}$, where $\mathcal{S}$ and $\mathcal{O}$ denote the unobserved and observed continuous state space respectively, $\mathcal{A}$ is the action space, $\mathbf{P}:\mathcal{S}\times \mathcal{O}\times \mathcal{A} \to \Delta(\mathcal{S} \times \mathcal{O})$ is the unknown transitional kernel, $R: \mathcal{S}\times \mathcal{O} \times \mathcal{A} \to \mathbb{R}$ is a bounded reward function, and $\gamma \in [0,1)$ is the discounted factor that balances the immediate and future rewards. $\mathcal{O}$ can also be treated as the observation space in the classical POMDP, then the process $\mathcal{M}$ can be summarized as $\{S_t,O_t,A_t,R_t\}^T_{t=1}$, with $S_t$ and $O_t$ as unobserved and observed state variables, $A_t$ as the action, and $R_t$ as the reward.

The objective of policy learning is to search an optimal policy, \(\pi^*\), which maximizes the expected discounted sum of rewards, using batch data obtained from a behavior policy \(\pi^b\). We assume the batch data consists of $n$ i.i.d. trajectories, i.e.,
\vspace{-3mm}
\begin{equation*}
\mathcal{D}_n=\{\mathcal{D}^i\}^{n}_{i=1}= \{O_0^i,A_0^i,R_0^i,O_1^i, \ldots ,O_T^i,A_T^{i}, R_T^{i},O_{T+1}^{i}\}^n_{i=1},
\vspace{-3mm}
\end{equation*}
where the length of trajectory $T$ is assumed to be fixed for simplicity, and the information on unobserved state $S^i_t$ is not available. In this paper, we focus on scenarios where the behavior policy, \(\pi^b\), maps both unobserved and observed state spaces to the action space, that is, \(\pi^b: \mathcal{S}\times \mathcal{O} \to \mathcal{A}\). Meanwhile, the target optimal policy only depends on the observed state: \(\pi^*: \mathcal{O} \to \mathcal{A}\). For a given target policy $\pi: \mathcal{O} \to \mathcal{A}$ , its state-value function is denoted as
\begin{equation}
V^{\pi}(s,o)=\mathbb{E}_{\pi}\Big[\sum^{\infty}_{k=0}\gamma^{k}R_{t+k}\mid S_t=s, O_t=o\Big],
\vspace{-2mm}
\end{equation}
where \(\mathbb{E}_{\pi}\) denotes the expectation with respect to the distribution whose actions at each decision point $t$ follow policy \(\pi\). Our goal is to utilize the batch data $\mathcal{D}_n$ to find the optimal policy \(\pi^*\) which maximizes the target policy value defined as
\begin{equation}
  J(\pi^*) = \argmax_{\pi} \mathbb{E}[V^{\pi}(S_0,O_0)],
\end{equation}
with \(\mathbb{E}\) representing the expectation in accordance with the behavior policy. Due to the unobserved state $S_t$, standard policy learning approach based on the Bellman optimality equation yields biased estimates. Thus, we first introduce an identification result to estimate the policy value for any target policy \(\pi\) with the help of some proxy variables. Subsequently, we employ policy gradient techniques to find the optimal policy \(\pi^*\).

\vspace{-3mm}
\section{Methodology}

Inspired by the proximal causal inference framework introduced by \citet{tchetgen2020introduction}, we initially present the identification result for policy evaluation in Section \ref{identification}. Following this, its associated minimax estimator for any given target policy $\pi$ is discussed in Section \ref{estimation}. Building on these policy evaluation findings, we detail a policy-gradient based approach in Section \ref{optimization} to identify the optimal policy within a policy class by maximizing the estimated policy value.

\subsection{Identification Results}
\label{identification}

In this section, we present the identification result for confounded POMDP setting. The derived identification equation serves as a foundation for off-policy evaluation (OPE) in the presence of unmeasured confounding, while also ensures the existence of bridge functions.

Building on the proximal causal inference framework proposed by \citet{tchetgen2020introduction}, we further assume the observation of reward-inducing proxy variables, \(W_t\), that relate to the action \(A_t\) solely through $\{S_t,O_t\}$. In practical scenarios, \(W_t\) could represent environmental factors correlated with the outcome \(R_t\), but remain unaffected by \(A_t\). For example, in healthcare applications, $W_t$ might include the choice of doctors or hospitals administering the treatment, or it could consist of variables that are either not accessible for decision-making due to privacy concerns or that become available only after the treatment. As for family panel studies \citep{bruderl2023german}, $W_t$ can be selected as the variables related to housing conditions, working environments and educational backgrounds of family members, while the action can be defined as the time spent with family. A representative Directed Acyclic Graph (DAG) of this can be seen in the left panel Figure \ref{pomdp}. The observed data for the confounded POMDP then have the form of
$$
\mathcal{D}_n=\{\mathcal{D}^i\}^{n}_{i=1}= \{O_0^i,W_0^i,A_0^i,R_0^i,O_1^i, \ldots ,O_T^i,W_T^i,A_T^{i}, R_T^{i},O_{T+1}^{i},W_{T+1}^i\}^n_{i=1}.
$$

\vspace{-2mm}

\begin{figure}[b]
\centering
\begin{minipage}{0.45\textwidth}
\centering
\scalebox{0.6}{
\begin{tikzpicture}[
    ->, >=stealth', auto, node distance=3.3cm, thick,
    main node/.style={
        circle,
        draw,
        minimum size=1.7cm,
        font=\sffamily\small
    },
    split node/.style={
        circle split,
        draw,
        font=\sffamily\small,
        minimum size=1.7cm,
        align=center,
        append after command={
            \pgfextra{
                \begin{scope} 
                    \clip (\tikzlastnode.center) -- (\tikzlastnode.east)
                        arc[start angle=0, end angle=-180, radius=0.86cm]
                        -- (\tikzlastnode.center) -- cycle;
                    \fill[lightgray] (\tikzlastnode.center) circle [radius=1.7cm];
                \end{scope}
                \begin{scope} 
                    \clip (\tikzlastnode.center) -- (\tikzlastnode.east)
                        arc[start angle=0, end angle=180, radius=0.86cm]
                        -- (\tikzlastnode.center) -- cycle;
                    \fill[blue!30] (\tikzlastnode.center) circle [radius=1.7cm];
                \end{scope}
            }
        }
    },
    shadow node/.style={
        font=\sffamily\small
    }
]

  \node[split node] (O) {
     $O_t$
    \nodepart{lower} $S_t$
  };
  \node[main node, fill=orange!30] (W) [above of=O] {$W_t$};
  \node[main node, fill=green!30] (R) [right of=W] {$R_t$};
  \node[main node, fill=red!30] (A) [right of=O] {$A_t$};

  \node[split node] (O+) [right of=A]{
    $O_{t+1}$
    \nodepart{lower}
    $S_{t+1}$
  };
  \node[main node, fill=orange!30] (W+) [above of=O+] {$W_{t+1}$};
  \node[main node, fill=green!30] (R+) [right of=W+] {$R_{t+1}$};
  \node[main node, fill=red!30] (A+) [right of=O+] {$A_{t+1}$};


  \path[every node/.style={font=\sffamily\small}]
    (O) edge[bend left=-45] (O+)
    (O) edge (A)
    (W) edge[dashed] (R)
    (A) edge (R)
    (O) edge (W)
    (O) edge (R)
    (O+) edge (A+)
    (W+) edge[dashed] (R+)
    (A+) edge (R+)
    (O+) edge (W+)
    (O+) edge (R+)
    (A) edge  (O+)
    (A+) edge (R+);

  \draw[->] ([xshift=-0.75cm]O.west) -- (O);
  \draw[->] (A+) -- ([xshift=1cm]A+.east) ;
  \draw[->, bend left=-45] (O+) to ([xshift=1.5cm]A+.south east);
\end{tikzpicture}}
\end{minipage}
\hfill
\begin{minipage}{0.45\textwidth}
\centering
\scalebox{0.6}{
\begin{tikzpicture}[
    ->, >=stealth', auto, node distance=3.3cm, thick,
    main node/.style={
        circle,
        draw,
        minimum size=1.7cm,
        font=\sffamily\small
    },
    shadow node/.style={
        font=\sffamily\small
    }
]
  \node[main node, fill=gray!30] (S) {$S_t$};
  \node[main node, fill=blue!30] (O) [above of=S] {$O_t$};
  \node[main node, fill=green!30] (R) [right of=O] {$R_t$};
  \node[main node, fill=red!30] (A) [right of=S] {$A_t$};

  \node[main node, fill=gray!30] (S+) [right of=A]{$S_{t+1}$};
  \node[main node, fill=blue!30] (O+) [above of=S+] {$O_{t+1}$};
  \node[main node, fill=green!30] (R+) [right of=O+] {$R_{t+1}$};
  \node[main node, fill=red!30] (A+) [right of=S+] {$A_{t+1}$};


  \path[every node/.style={font=\sffamily\small}]
    (S) edge[bend left=-45] (S+)
    (S) edge (A)
    (S) edge (R)
    (S) edge (O)
    (A) edge (R)
    (A) edge (S+)
    (S+) edge (O+)
    (S+) edge (A+)
    (S+) edge (R+)
    (A+) edge (R+);

    \draw[->] ([xshift=-0.75cm]S.west) -- (S);
  \draw[->] (A+) -- ([xshift=1cm]A+.east) ;
  \draw[->, bend left=-45] (S+) to ([xshift=1.5cm]A+.south east);

\end{tikzpicture}}
\end{minipage}
\caption{\yl{DAG of the proposed confounded POMDP (left) and classical POMDP (right).}}
\label{pomdp}
\end{figure}
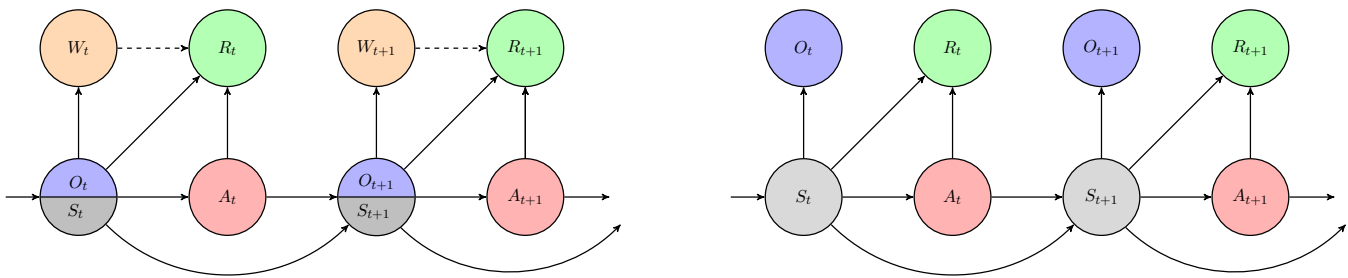

The left panel of Figure \ref{pomdp} can be viewed as a specific example of the proximal causal inference framework proposed by \citet{tchetgen2020introduction,cui2023semiparametric}. In our representation, we consider the previously observed state-action pair $(O_{t-1},A_{t-1})$ as the action-inducing proxy \(Z_{t}\). Therefore, all arrows related to \(Z_t\) from the original framework can be removed.


In contrast to the classical POMDP setting discussed in \citep{shi2022minimax,bennett2023proximal} and illustrated in the right panel of Figure~\ref{pomdp}, where the behavior policy depends solely on unobserved states, our proposed causal framework allows the behavior policy to be influenced by both observed and unobserved states with the help of additional reward-proxy variables $W_t$. Such modification more accurately captures real-world scenarios in which observable state variables can significantly affect decision-making during batch data collection process. A notable example is in precision medicine, where observable variables, such as laboratory results and the patient's current health status, often influence treatment choices.

Before presenting the identification results, we first formally introduce the basic assumptions of the confounded POMDP. Assumption \ref{markov} indicates that the future states are independent of the past given current full state and action $(S_t,O_t,A_t)$. Assumption \ref{reward} requires that the reward proxy $W_t$ relates to the unobserved state $S_t$ when conditioned on the observed state $O_t$, but not with the action $A_t$ when conditioned on the current full state $(S_t,O_t)$. Notably, Assumption \ref{reward} does not assume the causal relationship between $W_t$ and $R_t$, thus the dashed line between $W_t$ and $R_t$ in Figure \ref{pomdp}. Assumption \ref{action} requires that the previous observed state, $(O_{t-1},A_{t-1})$ does not influence the reward proxy $W_t$ and the reward $R_t$ given current full state and action. It can be easily verified that Assumptions \ref{markov}-\ref{action} are automatically satisfied by the DAG in Figure \ref{pomdp}.

\begin{asu}[Markovian] $(S_{t+1},O_{t+1})\indep \{S_j,O_j,A_j\}_{1\leq j < t}|(S_t,O_t,A_t)$, for $0\leq t \leq T$.
\label{markov}
\end{asu}
\begin{asu}[Reward Proxy]
$W_t \indep (A_t,S_{t-1},O_{t-1})|(S_{t},O_{t})$, and $W_t \noindep S_t|O_t $, for $1\leq t \leq T$.
\label{reward}
\vspace{-8.5mm}
\end{asu}
\begin{asu}[Action Proxy]
$(O_{t-1},A_{t-1})\indep (W_t,R_t)| S_t,O_t,A_t$ for $1\leq t \leq T$.
\label{action}
\end{asu}
However, we still cannot directly identify the value of target policy only with Assumptions \ref{markov}-\ref{action} by adjusting $(S_t,O_t)$, as $S_t$ is not observable. Thus, we also need the completeness assumption as stated in Assumption \ref{complete} to get around the unobserved state $S_t$.
\vspace{-2mm}
\begin{asu}[Completeness]
(a) For any square-integrable function $h$, $\mathbb{E}[h(S_t)|O_{t-1},A_{t-1},O_t,A_t]=0$ a.s. if and only if $h=0$ a.s. \newline
(b) For any square-integrable function $g$, $\mathbb{E}[g(O_{t-1},A_{t-1})|O_t,W_t,A_t]=0$ a.s. if and only if $g=0$ a.s.
\label{complete}
\vspace{-3mm}
\end{asu}
Completeness is a commonly made assumption in identification problems, such as instrumental variable identification \citep{newey2003instrumental,d2011completeness}, and proximal causal inference \citep{tchetgen2020introduction,cui2023semiparametric}. Assumption \ref{complete} (a) rules out conditional independence between $(O_{t-1},A_{t-1})$ and $S_t$ given $O_t$ and $A_t$, and indicates that the previous state-action pair $(O_{t-1},A_{t-1})$ should contain sufficient information from the unobserved state $S_t$. Assumption \ref{complete} (b) ensures the injectivity of the conditional expectation operator. Leveraging Picard’s Theorem \citep{kress1989linear}, the existence of bridge functions within a contextual bandit setting can be established \citep{miao2018identifying}.


Based on Assumptions \ref{markov}-\ref{complete}, we generalize the original result to an infinite horizon setting. Define the $Q$-bridge function and $V$-bridge function of the target policy $\pi$ as follows:
\begin{align}
\mathbb{E}[Q^{\pi}(O_t,W_t,A_t)|O_t,S_t,A_t] & = \mathbb{E}_{\pi}[\sum^{\infty}_{k=0}\gamma^kR_{t+k}|O_t,S_t,A_t], \label{qbridge}\\
\mathbb{E}[V^{\pi}(O_t,W_t)|O_t,S_t] & = \mathbb{E}_{\pi}[\sum^{\infty}_{k=0}\gamma^kR_{t+k}|O_t,S_t]. \label{vbridge}
\end{align}
Therefore, it it obvious that $V^{\pi}(O_t,W_t) = \int_{a\in \mathcal{A}}\pi(a|O_t)Q^{\pi}(O_t,W_t,a) da$. If there exists $V^{\pi}$ that satisfy \eqref{vbridge}, then the value of target policy $\pi$ can be identified by
\vspace{-3mm}
\begin{equation*}
  J(\pi)=\mathbb{E}[V^{\pi}(O_0,W_0)].
\vspace{-3mm}
\end{equation*}
Notice that bridge functions defined in \eqref{qbridge} and \eqref{vbridge} are not necessarily unique, but we can uniquely identify the policy value $J(\pi)$ based on any of them. We formally present the identification result in Theorem \ref{thm:identify}.
\begin{thm}
(Identification) For a confounded POMDP model whose variables satisfy Assumptions \ref{markov}-\ref{complete} and some regularity conditions, there always exist $Q$-bridges and $V$-bridges satisfying \eqref{qbridge} and \eqref{vbridge} respectively. Additionally, one particular $Q$-bridge and $V$-bridge can be obtained by solving the following equation
\begin{equation}
\mathbb{E}[Q^{\pi}(O_t,W_t,A_t)-R_t-\gamma V^{\pi}(O_{t+1},W_{t+1})|O_{t-1},A_{t-1},O_t,A_t]=0.
\label{identify}
\end{equation}
\label{thm:identify}
\end{thm}
\vspace{-12mm}
Theorem \ref{thm:identify} guarantees the existence of both $V$-bridges and $Q$-bridges. Additionally, the identification equation \eqref{identify} addresses the issue of the unobserved state $S_t$ by conditioning on previous state-action pair $(O_{t-1},A_{t-1})$, which forms the basis for estimating bridge functions and eventually estimate the policy value of the target policy $J(\pi)$.

\subsection{Minimax Estimation}
\label{estimation}
In this section, we discuss how to estimate the bridge functions using the pre-collected dataset $
\mathcal{D}_n=\{O_0^i,W_0^i,A_0^i,R_0^i,O_1^i, \ldots ,O_T^i,W_T^i,A_T^{i}, R_T^{i},O_{T+1}^{i},W_{T+1}^i\}^n_{i=1},
$ which consists of $n$ i.i.d copies of the observable trajectory $(O_t,W_t,A_t,R_t)^{T}_{t=1}$. Based on the identification equation \eqref{identify}, we have for the target policy $\pi$ and any function $f$,
$$
\mathbb{E}\Big[\Big(Q^{\pi}(O_t,W_t,A_t)-R_t-\gamma V^{\pi}(O_{t+1},W_{t+1})\Big)f(O_{t-1},A_{t-1},O_t,A_t)\Big]=0.
$$
We denote
$$
L_{\pi}(q,f) = \Big[q(O_t,W_t,A_t)-R_t-\gamma \int_{a\in \mathcal{A}}\pi(a|O_{t+1})q(O_{t+1},W_{t+1},a) da \Big]f(O_{t-1},A_{t-1},O_t,A_t),
$$
thus $\mathbb{E}[L_{\pi}(Q^{\pi},f)]=0$ for any $f \in \mathcal{F}$. Such observation directly leads us to the minimax estimator
\begin{equation}
\tilde Q^{\pi} = \argmin_{q\in \mathcal{Q}}\max_{f \in \mathcal{F}} \mathbb{E}[L_{\pi}(q,f)],
\label{qtilde}
\end{equation}
where we use the function class $\mathcal{Q}$ to model the $Q$-bridge function, the function class $\mathcal{F}$ to model the critic function $f$. The corresponding finite-sample estimator is then
\begin{equation}
\hat Q^{\pi} =\argmin_{q \in \mathcal{Q}}\max_{f\in \mathcal{F}}\mathbb{E}_{\mathcal{D}}[L_{\pi}(q,f)]+\lambda_n h_1^2(q)-\frac{1}{2}\mathbb{E}_{\mathcal{D}}[f^2]-\mu_n h^2_2(f),
\label{qhat}
\end{equation}
where $\mathbb{E}_{\mathcal{D}}$ denotes the sample average over all observed tuples $\{O_{t-1},A_{t-1},O_t,W_t,A_t,R_t,O_{t+1},W_{t+1}\}$, $h_1: \mathcal{Q} \to \mathbb{R}^+, h_2: \mathcal{F} \to \mathbb{R}^+$  are two regularizers and $\mu_n, \lambda_n$ are tuning parameters.

We observe that $\hat Q^{\pi}$ acts as a penalized estimator for $\tilde Q^{\pi}$. The term $\frac{1}{2}\mathbb{E}_{\mathcal{D}}[f^2]$ serves as an $L_2$ regularizer for the function class $\mathcal{F}$, which has been previously explored in the context of reinforcement learning literature \citep{antos2008learning, hoffman2011regularized}, as well as in the broader domain of minimax estimation problems \citep{dikkala2020minimax}. The component $\lambda_n h_1^2(q)$ aims to strike a balance between the model's fit regarding the estimated Bellman error and the complexity of the estimated $Q$-function. Similarly, $\mu_n h_2^2(f)$ is deployed to mitigate overfitting, especially when the function class $\mathcal{F}$ exhibits complexity. The estimated policy value can subsequently be calculated using the following equation,
\vspace{-2mm}
\begin{equation}
\hat J(\pi) = \mathbb{E}_{\mathcal{D}}\left[\int_{a\in \mathcal{A}}\pi(a|O_0)\hat Q^{\pi}(O_0,W_0,a) da\right].
\label{jhat}
\vspace{-2mm}
\end{equation}
The minimax optimization of \eqref{qhat} provides a clear direction to estimate the $Q$-bridge. In practice, we can use linear basis functions, neural networks, random forests and reproducing kernel Hilbert spaces (RKHSs), etc., to parameterize $q$ and $f$, and get the estimated $\hat Q^{\pi}$. However, directly solving the minimax optimization problem can be unstable due to its inherent complexities. Furthermore, representing $f$ within an arbitrary function class $\mathcal{F}$ poses additional intractability. Fortunately, we identify the  continuity invariance between the reward function and the optimal critic function $f^{*}(\cdot)$ in Theorem \ref{cts_critic}.

\begin{thm}
\label{cts_critic}
Suppose $\mathcal{F} \in L^2(C_0)$, and we define the optimal critic function as \newline $ f^*=\argmax_{f\in \mathcal{F}} \mathbb{E}[L_{\pi}(q,f)]$. Let $\mathbb{C}(\mathcal{O} \times \mathcal{A} \times \mathcal{O} \times \mathcal{A})$ be all continuous functions on $\mathcal{O} \times \mathcal{A} \times \mathcal{O} \times \mathcal{A}$. For any $(o^{-},a^-,o,a) \in \mathcal{O} \times \mathcal{A} \times \mathcal{O} \times \mathcal{A}$ and $s \in \mathcal{S}$, the optimal critic function $ f^*(O_{t-1},A_{t-1},O_t,A_t) \in   \mathcal{F}  \cap \mathbb{C}(\mathcal{O} \times \mathcal{A} \times \mathcal{O} \times \mathcal{A})$ is unique if the reward function $R(s,o,a)$ and the transition kernel $\mathbf{P}(s^+,o^+|s,o,a)$ are continuous over $(s,o,a)$, and the density of the target policy $\pi$ is continuous over $\mathcal{O} \times  \mathcal{A}$.
\end{thm}

Theorem \ref{cts_critic} demonstrates that, provided the reward function and the density of the target policy are continuous, which is widely held in real-world scenarios, the optimal \( f^*(\cdot)\) remains continuous. Meanwhile, for a positive definite kernel \(K\), a bounded RKHS denoted as
$ H_{\text{RKHS}}(C_0) := \{ f \in H_{\text{RKHS}} : \|f\|^2_{\mathcal{H}_{K}} \leq C_0 \} $
enjoys a diminishing approximation error to any continuous function class as \(C_0 \rightarrow \infty\) \citep{bach2017breaking}. Given this observation and the previously mentioned continuity invariance, we propose to represent the critic function within a bounded RKHS. We further show that by kernel representation, the original minimax optimization problem in \eqref{qtilde} can be decoupled into a single stage minimization problem in Theorem \ref{decouple}.


\begin{thm}
\label{decouple}
Suppose $\mathcal{F}$ belongs to a  bounded reproducing kernel Hilbert space (RKHS), i.e., $\mathcal{F} = \{f(o^-,a^-,o,a); f \in {\mathcal{H}_K},\|f\|^2_{\mathcal{H}_{K}}\leq C_0  \}$, then the original minimax problem defined in \eqref{qtilde} can be decoupled into the following minimization problem
\begin{align}
\label{loss}
& \tilde Q^{\pi} =\argmin_{q\in \mathcal{Q}} \mathbb{E}\Big[\Big(q(O_t,W_t,A_t)-R_t-\gamma \int_{a\in \mathcal{A}}\pi(a|O_{t+1})q(O_{t+1},W_{t+1},a) da \Big) \\
& K(\{O_{t-1},A_{t-1},O_t,A_t\}; \{\bar O_{t-1},\bar A_{t-1},\bar O_t, \bar A_t\}) \Big(q(\bar O_t,\bar W_t,\bar A_t)-\bar R_t-\gamma \int_{a\in \mathcal{A}}\pi(a|\bar O_{t+1})q(\bar O_{t+1},\bar W_{t+1},a) da \Big) \Big ] \nonumber,
\vspace{-3mm}
\end{align}
where $(\bar O_{t-1},\bar A_{t-1}, \bar O_t,\bar W_t,\bar A_t,\bar O_{t+1},\bar W_{t+1})$ is an independent copy of the transition pair \newline $(O_{t-1},A_{t-1}, O_t, W_t, A_t, O_{t+1},W_{t+1})$.
\vspace{-2mm}
\end{thm}

Theorem \ref{decouple} essentially transforms the minimax problem presented in \eqref{qtilde} into a single-stage minimization problem through its kernel representation, offering a direct approach to optimization. We note that methods have been developed to directly solve the sample version of \eqref{loss} in both MDP \citep{uehara2020minimax} and POMDP setting \citep{shi2022minimax}. However, in the context of batch reinforcement learning especially with continuous action space, optimization procedure may be unstable due to limited data without appropriate regularization. To address this, we additionally demonstrate that the finite-sample estimator, as formulated in \eqref{qhat}, also enjoys a closed-form solution for its inner-maximization part.
\begin{thm}
\label{finite_est}
Suppose  $\mathcal{F} = \{f(o^-,a^-,o,a); f \in {\mathcal{H}_K}  \}$, and $h_2^2(f) \vcentcolon =\|f\|^2_{\mathcal{H}_{K}}$ denotes the kernel norm of $f$, then the finite-sample minimax problem as shown in \eqref{qhat} can be decoupled into the following minimization problem
\vspace{-2mm}
\begin{equation}
\label{ustat}
\hat Q^{\pi}  \vcentcolon = \argmin_{q \in \mathcal{Q}} \mathcal{L}(q)  = \argmin_{q\in \mathcal{Q}} \Phi^{\top}(q)K_{nT}^{1/2}\Big[\frac{1}{2nT\mu_n}K_{nT}+I\Big]^{-1}K_{nT}^{1/2}\Phi(q)+\lambda_n h_1^2(q),
\end{equation}
where $K_{nT}=K\Big(\{O^i_{t-1},A^i_{t-1},O^i_t,A^i_t\}; \{O^{i'}_{t'-1}, A^{i'}_{t'-1}, O^{i'}_{t'}, A^{i'}_{t'}\}\Big)$ is the sample kernel matrix,  $\Phi(q)=[\delta^1_{1,\pi}(q),\delta^{1}_{2,\pi}(q),...,\delta^{1}_{T,\pi}(q),\delta^{2}_{1,\pi}(q),....,\delta^{n}_{T-1,\pi}(q),\delta^{n}_{T,\pi}(q)]^{\top} \in \mathbb{R}^{nT}$, and $\delta^{i}_{t,\pi}(q)=q(O^i_t,W^i_t,A^i_t)-R^i_t-\gamma \int_{a\in \mathcal{A}}\pi(a|O^i_{t+1})q(O^i_{t+1},W^i_{t+1},a) da$.
\vspace{-2mm}
\end{thm}
Notice that by including the kernel norm $h_2^2(f) \vcentcolon =\|f\|^2_{\mathcal{H}_{K}}$ as the penalty term, we drop the boundedness constraint on the function class $\mathcal{F}$ for the finite-sample estimator. Additionally, when selecting $\mu_n,\lambda_n$ such that $\mu_n,\lambda_n,\frac{1}{n\mu_n} \to 0$, estimating equation \eqref{ustat} converges to the form \eqref{loss}. Thus, equation \eqref{ustat} can be considered as a regularized variant  of \eqref{loss}, and provides a clear path in estimating the $Q$-bridge function. By parameterizing the \( Q \)-bridge function with parameter \( \theta \) using tools such as linear basis functions, RKHS, neural networks among others, we introduce an SGD-based algorithm as outlined in Algorithm \ref{SGD Algorithm} to determine the estimated \( \hat Q^{\pi} \).

\begin{algorithm}[t]
\setstretch{1}
	\caption{Off-Policy Evaluation}
	\label{SGD Algorithm}
	\begin{algorithmic}[1]
			\STATE \textbf{Input} observed transition pairs data $ \{O_{t-1},A_{t-1},O_t,W_t,A_t,R_t,O_{t+1},W_{t+1}\}^n_{t=1}$, target policy $\pi$.
			\STATE \textbf{Initialize} the  parameters of interests $\theta = \theta^{(0)}$, the mini-batch size $n_0$, the learning rate $\alpha_{0}$, the kernel bandwidth $\textit{bw}_0$, and the stopping criterion $\varepsilon$.
			\STATE \textbf{For iterations} $j=1$ to $k$
						\STATE \; \ Randomly sample a mini-batch $ \{O_{t-1},A_{t-1},O_t,W_t,A_t,R_t,O_{t+1},W_{t+1}\}^{n_0}_{t=1}$.
			          \STATE \;  \ Decay the learning rate $\alpha_{j} = \mathcal{O}(j^{-1/2})$.
			          \STATE \; \ Compute stochastic gradients with respect to $\theta$, $\nabla_{\theta}{\mathcal{L}(q)}$, in \eqref{ustat}.			
				\STATE \;  \ Update the parameters of interest as
					$\theta^{(j)} \leftarrow \theta^{(j-1)} - \alpha_{j}{\nabla}_{\theta}{\mathcal{L}(q)}.$
									\STATE \;  \ Stop if $\|\theta^{(j)} - \theta^{(j-1)} \| \leq \varepsilon$.
			  \STATE \textbf{Return}	 $\widehat \theta \leftarrow \theta^{(j)} $.
\end{algorithmic}
\end{algorithm}

\subsection{Policy Learning}
\label{optimization}
Section \ref{estimation} outlines the methodology for estimating the $Q$-bridge. Following this, the value of a specific target policy $\pi$ can be determined by $\mathbb{E}[V^{\pi}(O_0,W_0)] = \mathbb{E}\left[\int_{a \in \mathcal{A}} \pi(a|O_0)Q^{\pi}(O_0,W_0,a)da\right]$. It is important to note that the induced estimator $\hat Q^{\pi}$ is unbiased and exhibits a convergence rate of $O(n^{-1/2})$, as demonstrated in Section \ref{sec:theory}. Consequently, a natural approach for identifying the optimal policy $\pi^*$ is to search for the policy $\pi$ that maximizes the estimated policy value, i.e.
\begin{equation}
   \pi^* = \argmax_{\pi \in \Pi} J(\pi)= \argmax_{\pi \in \Pi} \mathbb{E}\left[\int_{a \in \mathcal{A}} \pi(a|O_0) Q^{\pi}(O_0,W_0,a)da\right].
\label{pi_star}
\end{equation}
However, this is a challenging problem due to the intractability of policy $\pi$, thus we propose to parameterize the policy distribution with $\pi( \zeta)$ to capture the learning process leading to the induced optimal policy, and the parameters corresponding to the optimal policy can then be obtained by solving the following optimization problem,
\begin{equation}
\hat \pi^*(o,a; \zeta) =  \argmax_{\pi \in \Pi} \hat J(\pi) =\argmax_{\pi \in \Pi}\mathbb{E}_{\mathcal{D}}\Big[\int_{a\in \mathcal{A}}\pi(a|O_0;\zeta)\hat Q^{\pi}(O_0,W_0,a;\theta)da\Big].
\label{pi_hat}
\end{equation}
Similar as $Q$-bridge, practical parameterization of distribution parameters, such as the mean and variance of the normal distribution or the parameters $(\alpha,\beta)$ for the beta distribution, can be achieved using linear basis functions, neural networks, and RKHSs.
\begin{algorithm}[t]
\setstretch{1}
	\caption{Off-Policy Learning}
	\label{policy_learn}
	\begin{algorithmic}[1]
			\STATE \textbf{Input} observed transition pairs data $ \{O_{t-1},A_{t-1},O_t,W_t,A_t,R_t,O_{t+1},W_{t+1}\}^n_{t=1}$.
			\STATE \textbf{Initialize} the  parameters of interests $\zeta=\zeta^{(0)}$, $\theta = \theta^{(0)}$, the mini-batch size $n_0$, the learning rate $\alpha_{0},\beta_0$, the kernel bandwidth $\textit{bw}_0$, and the stopping criterion $\varepsilon$.
			\STATE  \textbf{For iterations} $j=1$ to $k$
						\STATE \; \ Implement Algorithm \ref{SGD Algorithm} to get $\theta^{(j)}$.
			\STATE \; \ Compute the gradient with respect to $\zeta$, $\nabla_{\zeta}L_{\pi}(\zeta^{(j-1)},\theta^{(j)})$ in \eqref{pol_grad}.
              \STATE \;  \ Decay the learning rate $\beta_{j} = \mathcal{O}(j^{-1/2})$.
   		   \STATE \; \ Update the parameter $\zeta^{(j)} \leftarrow \zeta^{(j-1)} +\beta_{j}{\nabla}_{\zeta}\widehat{L_{\pi}}$.
			  \STATE \textbf{Return}	 $\widehat \zeta \leftarrow \zeta^{(j)} $.
\end{algorithmic}
\label{opl}
\end{algorithm}

To effectively tackle the optimization problem defined in \eqref{pi_hat}, we propose an algorithm that facilitates iterative updates to the parameters $(\zeta,\theta)$. Notice that for a predetermined $\zeta$, the corresponding $Q$-bridge $Q^{\pi(\zeta)}(o,w,a;\theta)$ can be fully determined by $\zeta$, thus $\theta$ can be considered as a function of $\zeta$. Consequently, to solve the optimization problem in  \eqref{pi_hat}, we start with an initial policy parameter $\zeta^{(0)}$, thereby defining the policy $\pi(\zeta^{(0)})$. Following the specification of initial value, we apply Algorithm \ref{SGD Algorithm} to estimate $\theta^{(0)}$ and the associated $Q$-bridge function $\hat Q^{\pi(\zeta^{(0)})}(O,W,A; \theta(\zeta^{(0)}))$.  Subsequently, the policy gradient with respect to $\zeta$ from \eqref{pi_hat} is calculated as:
\begin{align}
\label{pol_grad}
\nabla_{\zeta}L_{\pi}(\theta,\zeta) & =\mathbb{E}_{\mathcal{D}}[\nabla_{\zeta} \int_{a\in \mathcal{A}}\pi(a|O_0;\zeta)Q^{\pi}(O_0,W_0,a; \theta(\zeta))da], \\
& = \mathbb{E}_{\mathcal{D}}\Big[\int_{a\in \mathcal{A}}\Big\{\nabla_{\zeta} \pi(a|O_0;\zeta)Q^{\pi}(O_0,W_0,a; \theta(\zeta))+ \pi(a|O_0;\zeta)\nabla_{\zeta}Q^{\pi}(O_0,W_0,a; \theta(\zeta)) \Big\}da\Big]. \nonumber
\end{align}

Gradient ascent is employed to update $\zeta^{(0)}$, with the objective of obtaining a policy with a larger estimated value. This procedure is repeated to update the parameters $(\zeta,\theta)$ iteratively until convergence is attained, at which point $(\zeta,\theta)$ are considered to approximate the optimal policy and its corresponding $Q$-bridge function. This iterative update mechanism is outlined in Algorithm \ref{opl}. We note that this policy learning mechanism is broadly applicable to identification results based on different causal graphs. This can be achieved by parameterizing the policy class and replacing the policy evaluation step in Algorithm \ref{opl} with methods designed for alternative settings. This approach offers significant computational advantages when dealing with continuous action spaces, as traditional policy iteration algorithms typically require solving an additional optimization problem $\max_{a \in \mathcal{A}} Q^{\pi}(o, w, a)$ at each iteration.

\vspace{-3mm}
\section{Theoretical Results}
\label{sec:theory}

In this section, we first derive the global rate of convergence for $\hat Q^{\pi}$ in \eqref{qhat}, and the finite-sample error bound for the estimated policy value $\hat J(\pi)$ in \eqref{jhat}. We then extend the results to the estimated optimal policy, where we derive the regret bound for $\hat \pi^*$ defined in \eqref{pi_hat}. We note that while we use kernel representation to obtain the closed-form solution for the inner-maximization problem as demnstrated in \eqref{loss} and \eqref{ustat}, our theoretical results are based on the general form of \eqref{qhat}.  To simplify notation, we denote the $L_2$ norm with respect to the average state-action distribution in the trajectory $\mathcal{D}$ as $\|f\|^2 = \mathbb{E}\Big[\frac{1}{T}\sum^T_{t=1}f^2(\cdot)\Big]$, and the Bellman operator $\mathcal{T}_{\pi}$ with respect to the target policy $\pi$ as $\mathcal{T}_{\pi}(s,o,a;q) = \mathbb{E}\big [R_{t+1}+\gamma \int_{a\in \mathcal{A}}\pi(a'|O_{t+1})q(O_{t+1},W_{t+1},a') da  \mid S_t=s, O_t=o, A_t=a \big]$. Before presenting our main results, we first state several standard assumptions on policy class $\Pi$, reward function $R_t$, and function class $\mathcal{Q}$ and $\mathcal{F}$.
\begin{asu}
\label{polclass}
    The target policy class $\Pi=\{\pi_{\zeta}: \zeta\in Z \subset \mathbb{R}^p\}$, satisfies: \newline
(i) $H \subset \mathbb{R}^p$ is compact and let $\text{diam}(Z) = \sup\{\|\zeta_1 - \zeta_2\|_2 : \zeta_1, \zeta_2 \in H\}$. \newline
(ii) There exists $L_{Z} > 0$, such that for $\zeta_1, \zeta_2 \in H$ and for all $(o, a) \in \mathcal{O} \times \mathcal{A}$, the following holds
\vspace{-2mm}
\begin{equation*}
|\pi_{\zeta_1}(a|o) - \pi_{\zeta_2}(a|o)| \leq L_{Z} \|\zeta_1 - \zeta_2\|_2.
\vspace{-3mm}
\end{equation*}
\end{asu}
\begin{asu}
(i) The reward is uniformly bounded: $|R_t|\leq R_{\max}$ for all $t\geq 0$. \newline
(ii) The function class $\mathcal{Q}$ satisfies $\|q\|_{\infty}\leq q_{\max}$ for all $q \in \mathcal{Q}$, and  $Q^{\pi}\in \mathcal{Q}$ for all $\pi \in \Pi$.
\label{funcq}
\end{asu}
\begin{asu}
     The function class $\mathcal{F}$ satisfies (i) $0\in \mathcal{F}$; (ii)$\|f\|_{\infty}\leq f_{\max}$ for all $ f \in \mathcal{F}$; (iii) $\kappa = \inf \{ \| \tilde f\|: \|\mathcal{T}_{\pi}(s,o,a;q)-q(o,w,a) \|=1,q\in \mathcal{Q},\pi \in \Pi    \}>0$, where $\tilde f = \argmax_{f }\mathbb{E}[L_{\pi}(q,f)-\frac{1}{2} f^2]$.
\label{funcf}
\vspace{-9mm}
\end{asu}
Assumption \ref{polclass} is used to control the complexity of policy class, which are commonly assumed in policy learning problems \citep{liao2022batch,wang2022policy}. Assumption \ref{funcq} is a  regular assumptions to impose boundedness condition on reward and function class $\mathcal{Q}$ \citep{antos2008learning,bennett2021off}. Assumption \ref{funcf} similarly ensures the boundedness of function class $\mathcal{F}$. Additionally, the value of $\kappa$ measures how well the function class $\mathcal{F}$ approximates the Bellman error for all $q \in \mathcal{Q}, \pi \in \Pi$. This concept has been explored as the well-posedness condition within the field of off-policy evaluation \citep{chen2022well,miao2022off}. A strictly positive value of $\kappa$ indicates a substantial overlap between the behavior and target policies in terms of the state-action distribution, ensuring the identification of the true policy value $J(\pi)$.


\begin{asu}
\label{penalty}
(i) The regularization functionals, $h_1$ and $h_2$, are pseudo norms and induced by the inner products, $h_1(\cdot,\cdot)$ and $h_2(\cdot,\cdot)$, respectively. There exist constants, $C_1$ and $C_2$, such that $h_2(\tilde f(q))\leq C_1+C_2h_1(q)$ holds for all $q\in \mathcal{Q}$. \newline
(ii) Let $\mathcal{Q}_M = \{c+q: |c| \leq R_{\max},q\in \mathcal{Q},h_1(q)\leq M\}$,  and $\mathcal{F}_M = \{f: f\in \mathcal{F},h_2(f)\leq M\}$. There exists constant $C_3$ and $\alpha \in (0,1)$ such that for any $\epsilon, M>0$,
\vspace{-3mm}
\begin{equation*}
\max \left\{ \log N(\varepsilon, \mathcal{F}_M, ||\cdot||_\infty), \log N(\varepsilon, \mathcal{Q}_M, ||\cdot||_\infty) \right\} \leq C_3 \left( \frac{M}{\varepsilon} \right)^{2\alpha} .
\vspace{-3mm}
\end{equation*}
\end{asu}
Assumption \ref{penalty} characterizes the complexity of the function class $\mathcal{Q}$ and $\mathcal{F}$.  The condition that the regularizers be
pseudo-norms is satisfied for common function classes such as RKHS and Sobolev spaces \citep{geer2000empirical,thomas2016data}. The upper bound on $h_2$ is realistic when the transition kernel is sufficiently smooth \citep{farahm2016regularized}. We use a common $\alpha \in (0,1)$ for both $\mathcal{Q}$ and $\mathcal{F}$ to simplify the proof.

\begin{thm}
Suppose the target policy $\pi \in \Pi$, and let $\hat Q^{\pi}$ be the estimator defined in \eqref{qhat}. Suppose Assumptions  \ref{markov}-\ref{penalty} hold, and the tuning parameter satisfy $\frac{1}{\tau}n^{-\frac{1}{1+\alpha}} \leq \mu_n \leq \tau \lambda_n$ for some constant $\tau >0$. Then the following bounds hold with probability at least $1-\delta$ where $\delta \in (0,1)$,
\vspace{-3mm}
\begin{equation*}
\|\mathcal{T}_{\pi}(\cdot,\cdot,\cdot;\hat Q^{\pi})-\hat Q^{\pi}\|^2 \lesssim \frac{1}{\kappa^2}\Big[\lambda_n \left[ 1 +  h_1^2(Q^{\pi}) \right] \left\{ 1 + \log(1/\delta) \right\}] \Big],
\vspace{-3mm}
\end{equation*}
where the leading constant depend only on $(\tau,R_{\max},q_{\max},f_{\max},C_1,C_2,C_3,\alpha,\gamma)$.
\label{bell_err}
\end{thm}

In supplementary material, we have proved that $\|\hat Q^{\pi} - Q^{\pi}\|^2 \lesssim \|\mathcal{T}_{\pi}(\cdot,\cdot,\cdot;\hat Q^{\pi})-\hat Q^{\pi}\|^2$. Consequently, Theorem \ref{bell_err} demonstrates that $\hat Q^{\pi}$ serves as a consistent estimator for $Q^{\pi}$, provided that $\lambda_n = o_P(1)$. When the tuning parameter is selected such that $\lambda_n \asymp \mu_n$ and $\lambda_n \asymp n^{-1/(1+\alpha)}$, we achieve the optimal convergence rate of the Bellman error at $\hat Q^{\pi}$, which is $O_P(n^{-1/(1+\alpha)})$. Before delving into the finite-sample error bound for $\hat J(\pi)$, we first introduce some additional notations.

We define the discounted state-action visitation of target policy $\pi$ as $d^{\pi}(o,s,a)=(1-\gamma)\sum^{\infty}_{t=0}\gamma^{t}d^{\pi}_{t}(o,s,a)$, where $d^{\pi}_{t}$ is the density of the state-action pair at $t^{\text{th}}$ time point under the target policy $\pi$, and denote the average density over $T$ decision times under the behavior policy as $\bar d_T(o,s,a)$. We further define the direction function $e^{\pi}(o^-,a^-,o,a)$ by
\begin{equation}
\mathbb{E}[e^{\pi}(o^-,a^-,o,a)|s,o,a] = \frac{d^{\pi}(o,s,a)}{\bar d_T(o,s,a)}.
\end{equation}
The direction function $e^{\pi}$ is used to control the bias $|\hat J(\pi)-J(\pi)|$ caused by the penalization of $Q$-bridge in \eqref{qhat}. As demonstrated in \citet{uehara2020minimax, shi2022minimax}, the direction function $e^{\pi}$ enjoys the following property in our setting
\begin{align}
& \mathbb{E}\Big[\frac{1}{T}\sum^T_{t=1}e^{\pi}(O_{t-1},A_{t-1},O_t,A_t)\Big\{q(O_t,W_t,A_t)-\gamma \int_{a'\in \mathcal{A}} \pi(a'|O_{t+1})q(O_{t+1},W_{t+1},a')da'\Big\}\mid S_t,O_t,A_t\Big] \nonumber\\
& = -(1-\gamma)\mathbb{E}\Big[\frac{1}{T}\sum^T_{t=1} d^{\pi}_0(O_t,S_t,A_t)q(O_t,W_t,A_t)\Big].
\label{ortho}
\end{align}
Next, we define $\mathbb{E}[\xi^{\pi}(O_t,W_t,A_t)|S_t,O_t,A_t]=\mathbb{E}[\sum^{\infty}_{t=0} \gamma^{t} \frac{d^{\pi}(O_t,S_t,A_t)}{\bar d_T(O_t,S_t,A_t)}]$. Note that $\xi^{\pi}$ has a similar structure to the $Q$-bridge function, where the ``reward'' at time $t$ is defined as $\frac{d^{\pi}(O_t,S_t,A_t)}{\bar d_T(O_t,S_t,A_t)}$. Similar to the $Q$-bridge, $\xi^{\pi}$ satisfies the following Bellman-like equation as demonstrated in Theorem \ref{identification}:
\begin{align*}
\mathbb{E} & \Big[\xi^{\pi}(O_t,W_t,A_t)- e^{\pi}(O_{t-1},A_{t-1},O_t,A_t)+\\
& \qquad \qquad \qquad  \gamma\int_{a'\in \mathcal{A}}\pi(a'|O_{t+1})\xi^{\pi}(O_{t+1},W_{t+1},a')da' \mid O_{t-1},A_{t-1},O_t,A_t\Big]=0.
\end{align*}
We make the following smoothness assumption about $e^{\pi}$ and $\xi^{\pi}$.
\begin{asu}
The direction function $e^{\pi} \in \mathcal{F}$, and $\xi^{\pi} \in \mathcal{Q}$ for all $\pi \in \Pi$.
\label{direct}
\vspace{-3mm}
\end{asu}
The condition on direction function $e^{\pi}$ is to ensure that $e^{\pi}$ is sufficiently smooth, which will be used to show that the bias of the estimator in \eqref{qhat} decreases sufficiently fast to 0. The assumption on $\xi^{\pi}$ is analog to the assumptions used in partially linear regression literature \citep{geer2000empirical}. The counterpart of $\xi^{\pi} \in \mathcal{Q}$ in partially linear regression problem, $Y=g(Z)+X^{\top}\beta+\epsilon$ is the standard assumption that $\mathbb{E}[X|Z = \cdot]\in \mathcal{G}$, where $\mathcal{G}$ is the function class to model $g(Z)$. The last assumption concerns the coverage of the collected dataset.


\begin{asu}
(a) There exist positive constants $p_{1,\min},p_{2,\min}$ and $p_{1,\max}$, such that the visitation density $\bar d_T,d^{\pi}_0 (o,s,a)$ satisfy $p_{1,\min}\leq \bar d_T,d^{\pi}_0 \leq p_{1,\max}$ for all $\pi \in \Pi$, and the behavior policy $\pi^b\geq p_{2,\min}$ for every $(o,s,a) \in \mathcal{O}\times \mathcal{S}\times \mathcal{A}$. \newline
(b) The target policy $\pi$ is absolute continuous with respect to behaviour policy $\pi^b$ for all $\pi \in \Pi$, and $\mathcal{P}^{\pi}(o',s',w',a'|o,s,w,a)\leq p_{2,\max}$ for some positive constant $p_{2,\max}$, where $\mathcal{P}^{\pi}$ denotes the 1-step visitation density induced by target policy $\pi$.
\label{asu:den}
\end{asu}

We define $p_{\min} = \min \{p_{1,\min}, p_{2,\min}\}, p_{\max}=\max \{p_{1,\max},p_{2,\max}\}$. Assumption \ref{asu:den} (a) is frequently referred as the coverage assumption in RL literature \citep{precup2000eligibility,kallus2020statistically}, which guarantees the collected offline data sufficiently covers the entire state-action space. Assumption \ref{asu:den} (b) imposes one mild condition on the target policy, which essentially states that the collected batch data is able to identify the true value of target policy. We now analyze the performance error between the finite-sample estimator and the true policy value.

\begin{thm}
\label{finite_err}
   Suppose the assumptions in Theorem \ref{bell_err} hold. Additionally, suppose Assumption \ref{direct} and \ref{asu:den} hold, and $\lambda_n=o(n^{-1/(1+\alpha)})$ . Then, with probability at least $1-\delta$ where $\delta \in (0,1)$, the estimation error from \eqref{qhat} is upper bounded by
$$
|\hat J(\pi)-J(\pi) | \leq \frac{2[(1+\gamma)q_{\max}+R_{\max}]f_{\max}}{1-\gamma}\sqrt{\frac{\log(2/\delta)}{2n}}+o_P(n^{-1/2}).
$$
\end{thm}
From Theorem \ref{finite_err}, it is evident that the proposed estimator, as defined in \eqref{qhat}, achieves a convergence rate of $O(n^{-1/2})$. This finite-sample error bound effectively extends the result from \cite{shi2022minimax} to encompass a wider range of function classes, while still preserving the optimal convergence rate of $O(n^{-1/2})$. Moreover, Theorem \ref{finite_err} not only details the convergence rate for a specified target policy $\pi$, but also sets the groundwork for extending this outcome to the finite-sample regret bound applicable to policy learning.

\begin{proposition}
Let $\hat Q^{\pi}$ be the estimator defined in \eqref{qhat}. Suppose Assumptions  \ref{markov}-\ref{penalty} hold, and the tuning parameter satisfy $\frac{1}{\tau}n^{-\frac{1}{1+\alpha}} \leq \mu_n \leq \tau \lambda_n$ for some constant $\tau >0$. Then the following bound holds with probability at least $1-\delta$ where $\delta \in (0,1)$, for all $\pi \in \Pi$,
\vspace{-3mm}
\begin{equation*}
\|\mathcal{T}_{\pi}(\cdot,\cdot,\cdot;\hat Q^{\pi})-\hat Q^{\pi}\|^2 \lesssim \frac{p}{\kappa^2}\Big[\lambda_n \left[ 1 +  h_1^2(Q^{\pi}) \right] \left\{ 1 + \log(1/\delta) \right\}] \Big],
\vspace{-3mm}
\end{equation*}
where the leading constant depend only on $(\tau,R_{\max},q_{\max},f_{\max},C_1,C_2,C_3,\text{diam}(Z),L_Z,\alpha,\gamma)$.
\label{uni_err}
\end{proposition}

Proposition \ref{uni_err} extends the error bound in Theorem \ref{bell_err} to all $\pi \in \Pi$ with additional efforts in controlling the complexity of policy class $\Pi$. This demonstrates that $\hat Q^{\pi}$ is a consistent estimator for any $\pi$ in $\Pi$. Building on this, we analyze the suboptimality of the estimated optimal policy $\hat \pi^*$.

\begin{thm}
\label{optimality}
Suppose $\hat \pi^*$ and $\pi^*$ are defined in \eqref{pi_star} and \eqref{pi_hat} respectively, and the assumptions in Theorem \ref{bell_err} hold. Then with probability at least $1-\delta$ where $\delta \in (0,1)$, we have
\vspace{-3mm}
$$
J(\pi^*)-J(\hat \pi^*) \leq C(\delta)(p^{1/2}n^{-1/2})+o_P(p^{1/2}n^{-1/2}),
\vspace{-3mm}
$$
where $C(\delta)$ is a function of $(\tau,R_{\max},q_{\max},f_{\max},C_1,C_2,C_3,\text{diam}(Z),L_Z,\alpha,\gamma)$.
\end{thm}
Theorem~\ref{optimality} is based on the critical observation that
$ J(\pi^*) - J(\hat \pi^*) \leq (J(\pi^*) - \hat{J}(\pi^*)) + (\hat{J}(\pi^*) - \hat{J}(\hat \pi^*)) + (\hat{J}(\hat \pi^*) - J(\hat \pi^*))$,
where the error bounds for the first and last terms can be established from the uniform error bound on \( |\hat{J}(\pi) - J(\pi)| \) for all \(\pi \in \Pi\). Recall that \( p \) is the number of parameters in the policy, \( n \) is the number of i.i.d. trajectories in the collected dataset, and the regret of the estimated optimal policy is \( O(p^{1/2}n^{-1/2}) \). The first term represents the regret of an estimated policy as if the \( q \)-bridge were known beforehand, and the second term arises from the estimation error of the \( q \)-bridge function. Theorem~\ref{optimality} essentially characterizes the regret of the estimated optimal in-class policy without double robustness, achieving the optimal minimax convergence rate of \( O(n^{-1/2}) \).

\vspace{-3mm}
\section{Simulation Studies}
\label{sec:results}
In this section, we evaluate our proposed method through extensive simulation studies. Since our framework introduces the reward-inducing proxy variable and is designed for both continuous state and action spaces, existing methods for POMDPs are not directly applicable \citep{shi2022minimax, miao2022off}. Therefore, we compare our proposed method to other methods designed for MDP scenarios. For policy evaluation, we compare our method to the MDP version of our proposed approach, as well as an augmented MDP, hereafter referred to as MDPW, with $(O_t, W_t)$ as the new observed state variables. Specifically, we adopted the method by \citet{uehara2020minimax}, treating \(O_t\) (MDP) and $(O_t,W_t)$ (MDPW) as the state variables as two baselines.. For policy learning, we benchmark our method against state-of-the-art baselines, including implicit Q-learning (IQL) \citep{kostrikov2021offline}, soft actor-critic (SAC) \citep{haarnoja2018soft} and conservative Q-learning (CQL) \citep{kumar2020conservative}.

For practical implementation, both \( Q^{ \pi}(\cdot; \theta) \) and \( \pi(\cdot; \zeta) \) need to be parameterized. We propose to parameterize \( Q^{\pi} \) as a RKHS using second-degree polynomial kernels. The critic function class $\mathcal{F}$ is specified as a Gaussian RKHS with bandwidths selected by the median heuristic \citep{fukumizu2009kernel}.
The penalty terms $h_1(q), h_2(f)$ are selected as the kernel norm of $q,f$ respectively. Thus, it can be verified that Assumptions \ref{polclass}-\ref{penalty} are automatically satisfied by the above choice.


To determine the values of the tuning parameters \( (\lambda_n, \mu_n) \), we employ a k-fold cross-validation approach. For each candidate pair of tuning parameters, we implement Algorithm \ref{opl} to search for the optimal policy on the training set and then evaluate the learned optimal policy on the validation set. We select the optimal \( (\lambda_n, \mu_n) \) that maximizes the estimated optimal policy value on the validation set, given by: $(\hat \lambda_n, \hat \mu_n) = \arg\max_{\lambda_n, \mu_n} \frac{1}{k} \sum_{r=1}^k \hat J_{\lambda_n, \mu_n}^{(r)}(\hat \pi^*)$, where \( \hat J_{\lambda_n, \mu_n}^{(r)}(\hat \pi^*) \) represents the estimated optimal policy value on the \( r \)-th validation set. We conduct numerical experiments on a synthetic environment to evaluate the finite sample performance of our proposed method. Both state and action space are continuous for the synthetic environment, and the discount factor is specified as $\gamma=0.9$ for all experiments.


We consider a one-dimensional continuous state space and a continuous action space for the synthetic environment, where the unobserved initial state follows $S_0 \sim \text{Unif}(-0.5, 0.5)$. The observed state $O_t$ is generated according to the additive noise model, i.e., $O_t = S_t + \mathcal{N}(0, 1)$. The reward proxy, reward function, and state transition are given by
\begin{gather*}
\vspace{-3mm}
    R_t = A_t(O_t - 0.2W_t - 0.8S_t) - 0.8A_t^2, \qquad W_t = S_t - 0.5O_t + N(0, 0.3^2),
    \\S_{t+1} = 0.8O_t - 0.3A_t + N(0, 0.1^2).
\vspace{-3mm}
\end{gather*}
To ensure sufficient coverage of the observed dataset, we set the behavior policy as  $A_t \mid S_t, O_t \sim \text{clip}(N(-S_t/3, 0.4^2), -1, 1)$ to generate the batch dataset.

We first conduct policy evaluation by selecting two different target policies. The first is designed to be similar to the behavior policy, while the second is near-optimal and therefore substantially different from the behavior policy. We then apply Algorithm \ref{SGD Algorithm} to estimate the target policy value, and consider different sample sizes for the observed dataset in which $(n,T)=\{(25,25),(50,25),(75,50),(100,50)\}$. To evaluate the model performance, we generate 1,000 independent trajectories under the target policy and define the true policy value as the simulated mean cumulative reward. We compare the MSE of our proposed estimator with its MDP and MDPW counterparts. The simulation results are shown in Figure \ref{fig:ope_env1}.
\begin{figure}[t]
    \centering
    \includegraphics[width=0.9\linewidth]{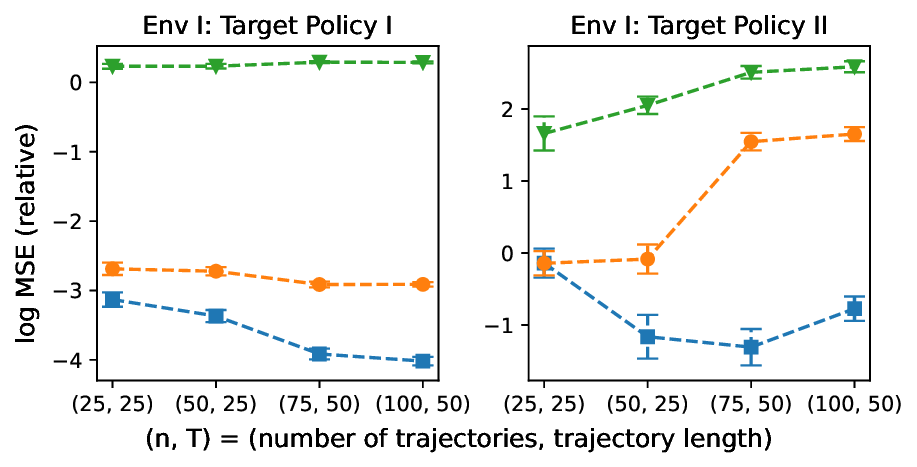}
    \caption{Logarithms of relative MSEs of the proposed (blue sqaures), MDPW (orange circles), and MDP (green triangles) estimators and their associated 95\% confidence interval, based on 50 simulations, with different choices of the sample size and length of trajectory $(n, T)$. }
    \label{fig:ope_env1}
\end{figure}

Figure \ref{fig:ope_env1} illustrates that our proposed method consistently achieves lower bias in policy evaluation across all settings. This improvement stems from leveraging the identification result in Theorem \ref{identification}, which enables the recovery of lost information from unobserved variables by using reward-inducing proxies. In contrast, MDP-based methods typically exhibit higher bias and variance because they treat the observed state as the true state, even when the observed state is at best a noisy representation of the true state. Thus, these methods could easily lead to a biased estimate by ignoring the impact of unmeasured confounders.

For policy learning, we consider both Gaussian and Beta distribution policy class, and parameterize distribution parameters using either a linear basis or neural network. For Beta distribution policy class with a linear basis, we define
$\pi^{\text{Linear}}_\zeta(a \mid o) \sim 2Beta(\log(1+\exp(\zeta_1^To)), \log(1+\exp(\zeta_2^To)))-1$, where $o$ is the state vector including an intercept, and $\zeta_1, \zeta_2 \in \mathbb{R}^d$ with the same dimension as $o$. For the neural network parameterization, we consider a one‐hidden‐layer MLP to approximate the $(\alpha, \beta)$ in the Beta distributions. For Gaussian distribution policy class, the policy parameters are similarly parameterized by $\pi(a \mid o) \sim \tanh\left(\mathcal{N}(\mu(o), \sigma^2(o))\right)$, where both the mean and standard deviation are approximated by either the same linear basis or neural network.  While the neural network policy parameterization is capable of learning more complex optimal policies, the linear basis provides better interpretability and are often preferable in real-world applications.

To evaluate the performance of our proposed method, we apply Algorithm~\ref{opl} and compare its results against the MDP and MDPW variants of the same approach, as well as against state-of-the-art policy learning algorithms including IQL, SAC, and CQL. To ensure a fair comparison, we employ the identical policy class for all policy parameterizations of the MDP and MDPW counterparts of our proposed method. Since IQL, SAC, and CQL are designed for deep reinforcement learning, we restrict their policy class to a Gaussian neural network. All data are generated using the synthetic environments and behavior policies described in the policy evaluation section, and Table~\ref{tab:opl} summarizes the averaged results over 50 simulation runs.

Table \ref{tab:opl} indicates that our proposed method consistently outperforms competing approaches across varying sample sizes while maintaining comparable variance. This advantage is primarily due to the unbiased estimation of policy value at each iteration, which results in accurate policy gradients and ensures that policies are updated in the correct direction. In contrast, the MDP and MDPW variants of our method may suffer from biased estimation of the target policies, causing suboptimal policy updates and therefore poorer performance. Furthermore, methods originally designed for MDP settings (i.e., IQL, SAC, and CQL) directly solve the Bellman optimality equation, which does not hold in the presence of unmeasured confounders. As a result, these methods perform worse when unmeasured confounders are significant. Lastly, our proposed method achieves stable performance even with relatively small sample sizes, which is particularly desired in real-world applications where data can be limited.

\begin{table}[t]
    \centering
    \footnotesize
    \begin{tabular}{l|c|c|c|c}
        \hline
         Method/(n, T) & (25, 25) & (50, 25) & (75, 50) & (100, 50)\\
         \hline
         Proposed-NN-Gaussian	&	1.71 (0.43)	&	1.80 (0.19)	&	1.80 (0.16)	&	1.77 (0.13)	\\
Proposed-NN-Beta	&	1.62 (0.36)	&	1.68 (0.24)	&	1.71 (0.13)	&	1.72 (0.12)	\\
Proposed-Linear-Gaussian	&	1.66 (0.13)	&	1.65 (0.09)	&	1.68 (0.07)	&	1.71 (0.07)	\\
Proposed-Linear-Beta	&	1.53 (0.11)	&	1.55 (0.10)	&	1.55 (0.09)	&	1.58 (0.10)	\\
         \hline
MDPW-NN-Gaussian	&	1.44 (0.29)	&	1.41 (0.23)	&	1.41 (0.20)	&	1.33 (0.36)	\\
MDPW-NN-Beta	&	1.27 (0.38)	&	1.34 (0.31)	&	0.93 (0.43)	&	0.95 (0.33)	\\
MDPW-Linear-Gaussian	&	1.41 (0.19)	&	1.42 (0.12)	&	1.41 (0.10)	&	1.37 (0.10)	\\
MDPW-Linear-Beta	&	1.27 (0.14)	&	1.28 (0.13)	&	1.02 (0.58)	&	0.98 (0.51)	\\
MDPW-SAC	&	0.72 (1.13)	&	1.04 (0.34)	&	1.23 (0.17)	&	1.26 (0.15)	\\
MDPW-CQL	&	1.41 (0.39)	&	1.44 (0.39)	&	1.30 (0.32)	&	1.29 (0.30)	\\
MDPW-IQL	&	0.61 (0.33)	&	0.90 (0.21)	&	0.94 (0.16)	&	0.98 (0.13)	\\
\hline
MDP-NN-Gaussian	&	0.55 (0.39)	&	0.10 (0.28)	&	-0.28 (0.10)	&	-0.18 (0.24)	\\
MDP-NN-Beta	&	-0.21 (0.19)	&	-0.32 (0.10)	&	-0.24 (0.11)	&	-0.22 (0.17)	\\
MDP-Linear-Gaussian	&	0.86 (0.53)	&	0.91 (0.48)	&	1.08 (0.14)	&	1.04 (0.12)	\\
MDP-Linear-Beta	&	0.46 (0.19)	&	0.38 (0.09)	&	0.11 (0.15)	&	0.10 (0.21)	\\
MDP-SAC	&	0.22 (0.64)	&	0.31 (0.34)	&	0.28 (0.12)	&	0.32 (0.10)	\\
MDP-CQL	&	0.44 (0.35)	&	0.50 (0.27)	&	0.41 (0.17)	&	0.36 (0.18)	\\
MDP-IQL	&	0.29 (0.35)	&	0.57 (0.25)	&	0.66 (0.19)	&	0.73 (0.14)	\\
         \hline
    \end{tabular}
    \caption{The mean and standard deviation (in parentheses) of the learned optimal policy value over 50 simulation runs. The true policy value are obtained from Monte Carlo simulation with $T = 100, n=1000$.}
    \label{tab:opl}
\end{table}
In terms of the comparison between the neural network and linear policy classes, neural networks show better performance due to the richness of the function class, but the linear basis is capable of achieving comparable performance with fewer and more interpretable parameters, and thus may be of more interest in real-world applications, as demonstrated in Section \ref{sec:pairfam}.

\section{Application to Pairfam Dataset}
\label{sec:pairfam}
We applied our proposed method to the Panel Analysis of Intimate Relationships and Family Dynamics (Pairfam) dataset \citep{bruderl2023german}. Initiated in 2008, Pairfam is a comprehensive longitudinal dataset designed to explore the evolution of romantic relationships and family structures within Germany. For this study, we utilized the most recent release of the Pairfam dataset, which encompasses survey results from 14 waves \citep{bruderl2023german}. Given that the true feelings and thoughts of participants regarding their relationships are unobserved and survey responses are proxies for these sentiments, we treated the evolution of relationships as a confounded Partially Observable Markov Decision Process (POMDP). Our objective was to use the Pairfam dataset to estimate the optimal policy for maximizing long-term satisfaction in romantic relationships.

We define the immediate reward, $R_t$, as the mean relationship satisfaction reported by the couple during each survey wave. The continuous action variable, $A_t$, is selected as the frequency of sharing private feelings and communicating with partners, referred to as intimacy frequency. The observed state variables at each time point, $O_t$, are represented by a 6-dimensional vector, which includes frequency of conflicts, frequency of appreciation, health status, future orientation of the couple, sexual satisfaction, and satisfaction with friendship of the couple. We defined the reward-proxy, $W_t$, as a 4-dimensional vector that includes housing condition, household income, division of labor between the couple, and presence of relatives during the interview. Based on romantic relationship literature, all variables in $W_t$ could be considered as independent of $A_t$ when conditioned on observed states.
To ensure data quality, we exclude samples with more than three missing values for any variable and applied matrix completion methods \citep{van2011mice} to impute the remaining missing data. Consequently, our refined dataset includes 809 trajectories ($n = 809$) each with 14 time points ($T = 14$).

To conduct policy learning on the dataset, we consider both neural network and linear policy class as discussed in Section \ref{sec:results}. We choose the same network structure as in Section \ref{sec:results}, and the linear policy class is defined as $\pi_{\zeta}(a \mid o) \sim \text{Beta}(1 + \zeta_1\text{expit}(\zeta_2^\top o), 1 + \zeta_3\text{expit}(\zeta_4^\top o))$, where $\zeta = [\zeta_1, \zeta_2,\zeta_3,\zeta_4]^\top$ and $o$ is a 7-dimensional vector that includes the intercept along with the 6 observed state variables discussed previously. For numerical stability, we constrain $\zeta_1,\zeta_3 \in (0,20)$. The forms $1 + \zeta_i\text{expit}(\zeta_j^\top o)$ for $i = 1, 3, j=2, 4$ are specifically selected for normalization purposes, enabling us to constrain the parameters associated with the beta distribution to be in the range $(1,21)$, thus avoiding extremely large or small values that may result in abnormal policies.

We set the discount factor at $\gamma$ = 0.9 and employ our proposed method to identify the optimal policy that maximizes long-term relationship satisfaction. To ensure robustness, each simulation randomly selects 400 trajectories as the training data and uses the remaining trajectories for testing data. Following \citet{luckett2019estimating}, we use a Monte Carlo approximation of the policy value to evaluate performance of each method. Specifically, we train the policy on the training data and then apply our proposed policy evaluation method from Algorithm~\ref{SGD Algorithm} to the testing set to evaluate the learned optimal policy, obtaining an estimate of the learned policy’s value. As shown in Figure~\ref{fig:ope_env1}, our proposed OPE method exhibits the smallest bias in the presence of unmeasured confounders compared to alternative methods.

Figure \ref{bootstrap} presents boxplots of the estimated policy values on both the training and testing sets for each method based on 100 simulation runs, with the baseline representing the observed discounted return. As shown in Figure \ref{bootstrap}, the proposed method achieves the best performance in terms of the improvement on policy value by taking unobserved confounders into account, which is consistent with the results presented in Section \ref{sec:results}. In contrast, the MDP-based method tends to overestimate the policy value, which may be undesirable in safe-critical scenarios. Additionally, we observe that the linear policy class shows similar performance compared to the neural network policy class. Since the linear basis is more interpretable in such a setting, we further examine the learned optimal policy obtained using the linear policy class.

\begin{table}[b]
\centering
\begin{tabular}{c|ccccccc}
\hline
 & Intercept & Conflicts & Appreciation & Health & Commitment & Sex & Friendship \\
\hline
$\zeta_2$ & -0.0626 & -1.1602 & 1.4199 & 1.3774 & 1.1408 & 0.9960 & -0.2873 \\
\hline
$\zeta_4$ & -0.8386 & 1.4781 & 0.1855 & -0.0315 & -0.7316 & 0.3191 & -0.6859 \\
\hline
\end{tabular}
\caption{Estimated Coefficients of the Optimal Policy.}
\label{coef}
\end{table}

\begin{figure}[t]
   \begin{minipage}{0.58\textwidth}
     \centering
     \includegraphics[width=0.95\linewidth]{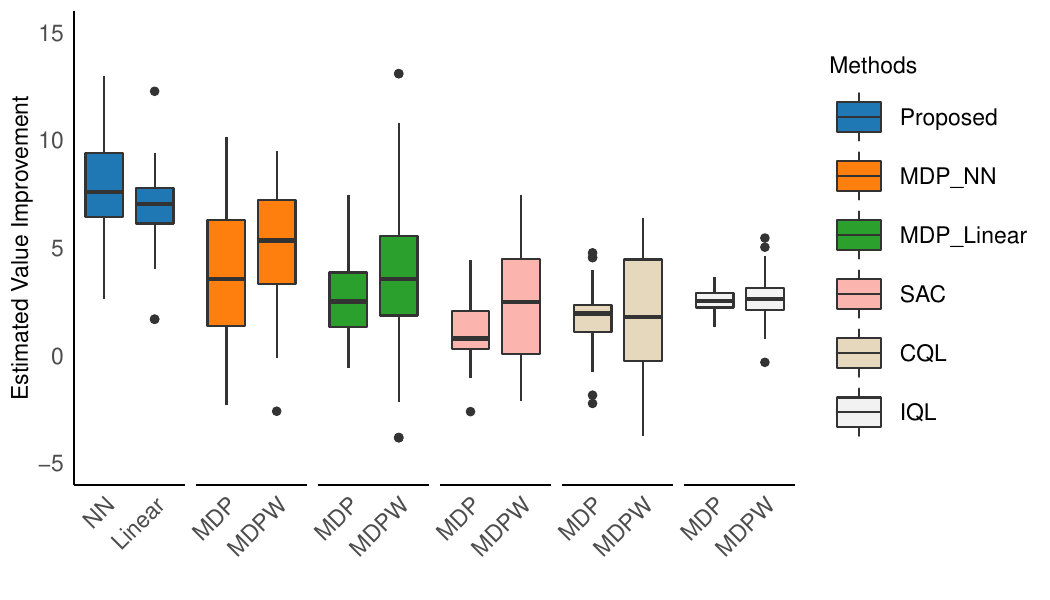}
     \caption{Boxplots of discounted reward improvements under 100 simulation runs with $\gamma=0.9$. The first two boxes represent the proposed method using neural network and linear policy class. The remaining boxes show the MDP and MDPW version of competing methods.}\label{bootstrap}
   \end{minipage}\hfill
   \begin{minipage}{0.37\textwidth}
     \centering
     \includegraphics[width=0.95\linewidth]{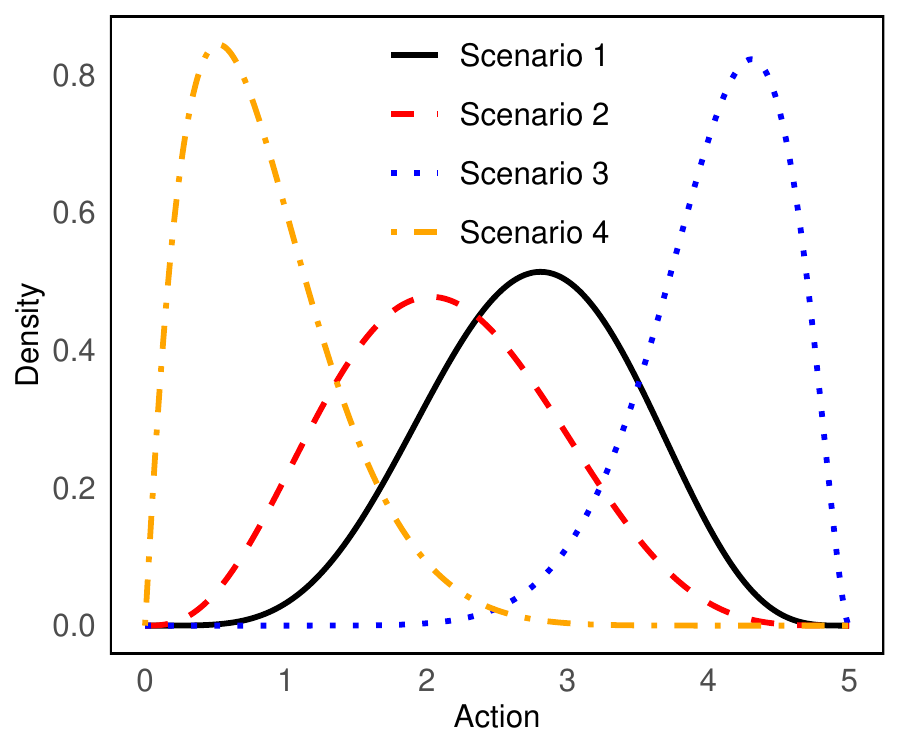}
\vspace{-3.5mm}
     \caption{The estimated optimal policy distribution under typical states. The corresponding states are defined in Table \ref{scenarios}.}\label{pairfam_pol}
   \end{minipage}
\end{figure}

For linear policy class, the estimated values of $\zeta = [\zeta_1, \zeta_2, \zeta_3, \zeta_4]^\top$ are shown in Table \ref{coef}, with $\zeta_1 = 12.3463$ and $\zeta_3 = 7.4619$. To better illustrate the learned policy, we demonstrate the learned optimal policy under four different observed states. In romantic relationship studies, researchers have classified couples into different categories based on their commitment levels and couple dynamics (e.g., negative interactions) \citep{beckmeyer2021identifying}. Thus, we chose commitment and conflict frequency as a proxy for couple dynamics. Consequently, we divided all the samples in the dataset into four categories based on whether their commitment and conflict levels are higher than the sample means, as shown in Table \ref{scenarios}. It is important to note that all variables have been normalized to have mean 0 and variance 1. The values of observed state variables, other than conflicts and commitment, are selected as the sample means of those variables within each category. Figure \ref{pairfam_pol} shows the induced optimal policy of each observed states.

\begin{table}[t]
\centering
\begin{tabular}{c|cccccc}
\hline
Scenarios & Conflicts & Appreciation & Health & Commitment & Sex & Friendship \\
\hline
Scenario 1 & 0.67 & -0.02 & 0.03 & 0.51 & 0.01 & 0.02 \\
\hline
Scenario 2 & -0.65 & -0.21 & -0.12 & -0.85 & -0.19 & -0.14 \\
\hline
Scenario 3 & -0.78 & 0.43 & 0.09 & 0.52 & 0.38 & 0.19 \\
\hline
Scenario 4 & 0.93 & -0.66 & -0.14 & -1.16 & -0.61 & -0.30 \\
\hline
\end{tabular}
\caption{Observed state variables for each scenario.}
\label{scenarios}
\end{table}

Figure \ref{pairfam_pol} suggests that for couples experiencing a high frequency of conflicts and low levels of commitment (Scenario 4), fewer instances of sharing feelings and thoughts should be undertaken to improve satisfaction. In contrast, more frequent sharing  is preferred for couples experiencing low conflicts and high commitment (Scenario 3). Additionally, moderate to high levels of sharing feelings and thoughts are preferred for couples with moderately high commitment and conflict levels (Scenario 1) and couples with moderately low commitment and conflict levels (Scenario 2). These results are consistent with the literature on the effects of sharing feelings and thoughts in romantic relationships. Although relationship researchers generally acknowledge the positive effects of sharing feelings (e.g., openness) \citep{ogolsky2013meta}, some studies have found that withholding feelings can be beneficial to relationship quality for individuals with high levels of social anxiety \citep{kashdan2007social} and for couples with a more communal orientation \citep{Le2013when}. Additionally, partner appreciation has been found to foster positive interactive dynamics through validation, thereby improving relationship satisfaction \citep{algoe2012find}.

\vspace{-3mm}
\section{Discussion}
In this paper, we proposed a novel framework to conduct offline policy evaluation and policy learning on continuous action space with the existence of unmeasured confounder. We extended the proximal causal inference framework \citep{cui2021semiparametric} to an infinite horizon setting for the identification of policy value for a fixed target policy, and developed the corresponding minimax estimator. Based on the proposed estimator, we further developed a policy-gradient based method to search for the in-class optimal policy. The PAC bound of the proposed algorithm is provided to analyze its sample complexity.

Several improvements and extensions are worth exploring in the future. First, we assumed there is no coverage issue for the batch dataset. However, this assumption may not hold in real-world scenarios, such as medical applications where sample sizes are generally small and data is expensive to obtain. Thus, properly addressing the data coverage issue under the POMDP setting for both policy evaluation \citep{zhang2024curses} and policy learning would be an interesting topic. Second, as demonstrated in real-data application, different observed state variables play various roles in formulating the policy. Meanwhile, conducting variable selection in an offline RL setting remains a challenging task, as there is no ground truth available for performance comparison. Systematically studying this problem would greatly improve the generalization of RL techniques. Finally, the proposed algorithm requires relatively large computation and memory resources due to the necessity of evaluating a new policy at each iteration. Therefore, developing a more efficient algorithm is desirable. A possible approach is to directly identify the policy gradient under the POMDP \citep{hong2023policy}, however, such identification in continuous action space remains a challenge.

\bibliographystyle{asa}
\bibliography{causal.bib}

\end{document}